%% file: TilingAlgoForChannels_hal.tex
\documentclass{article}[12pts]

\usepackage{setspace}
\usepackage{graphicx} 
\usepackage{subfigure} 

\usepackage{amssymb}
\usepackage{amsfonts,amsmath,amsthm}
\usepackage{color} 
 \usepackage{bbm} 
\usepackage{algpseudocode}

\newtheorem{theorem}{Theorem}
\newtheorem{assumption}{Assumption}

\newcommand{\eqdef}{\ensuremath{\stackrel{\mathrm{def}}{=}}}

\newcommand{\normi}[1]{{\left\Vert #1 \right\Vert}_{\infty}}
\newcommand{\argmax}{\operatornamewithlimits{argmax}}

\def\1{\mathbbm{1}}
\def\P{\mathbb{P}}
\def\E{\mathbb{E}}
\def\Xset{\ensuremath{\mathsf{X}}}

\def\Yset{\ensuremath{\mathsf{Y}}}

\def\Rset{\ensuremath{\mathbb{R}}}

\def\Aset{\mathsf{A}}

\def\eqsp{\;}
\def\eqsp{\;}
\newcommand{\CPE}[3][]
{\ifthenelse{\equal{#1}{}}{\operatorname{E}\left[\left. #2 \, \right| #3 \right]}{\operatorname{E}^{#1}\left[\left. #2 \, \right | #3 \right]}}
\newcommand{\CP}[3][]
{\ifthenelse{\equal{#1}{}}{\mathbb{P}\left[\left. #2 \, \right| #3 \right]}{\mathbb{P}^{#1}\left[\left. #2 \, \right | #3 \right]}}
\newcommand{\note}[2]{}

\title{Regret Bounds for Opportunistic Channel Access} 
\author{Sarah Filippi, Olivier Capp\'{e} and Aur\'{e}lien Garivier \\
LTCI, TELECOM ParisTech and CNRS, 46 rue Barrault, 75013 Paris, France\thanks{This work is partially supported by Orange Labs under contract n\textsuperscript{o}289365.}\\
(filippi, cappe, garivier)@telecom-paristech.fr
}
\date{}

\begin{document} 
\maketitle

\begin{abstract}
  We consider the task of opportunistic channel access in a primary system composed of independent
  Gilbert-Elliot channels where the secondary (or opportunistic) user does not dispose of a priori
  information regarding the statistical characteristics of the system. It is shown that this
  problem may be cast into the framework of model-based learning in a specific class of Partially
  Observed Markov Decision Processes (POMDPs) for which we introduce an algorithm aimed at striking
  an optimal tradeoff between the exploration (or estimation) and exploitation requirements. We
  provide finite horizon regret bounds for this algorithm as well as a numerical evaluation of its
  performance in the single channel model as well as in the case of stochastically identical
  channels.
\end{abstract}

\section{Introduction}
In recent years, opportunistic spectrum access for cognitive radio has been the focus of
significant research efforts
\cite{akyildiz:won-yeol:vuran:mohanty:2008,haykin:2005,mitola:2000}. These works propose to improve
spectral efficiency by making smarter use of the large portion of the frequency bands that remains
unused. In Licensed Band Cognitive Radio, the goal is to share the bands licensed to primary users
with non primary users called secondary users or cognitive users. These secondary users must
carefully 
identify available spectrum resources and communicate avoiding to disturb the primary
network. Opportunistic spectrum access thus has the potential for significantly increasing the
spectral efficiency of wireless networks.

In this paper, we focus on the opportunistic communication model previously considered by
\cite{Liu:Zhao:08,Zhao:al:07bis}, which consists of $N$ channels in which a single secondary user
searches for idle channels temporarily unused by primary users. The $N$ channels are modeled as
Gilbert-Elliot channels: at each time slot, a channel is either idle or occupied and the
availability of the channel evolves in a Markovian way. Assuming that the
secondary user can only sense $M \ll N$ channels simultaneously \cite{Lai:al:08,Liu:Zhao:08,Zhao:al:08}, his main task is to choose which
channel to sense at each time aiming to maximise its expected long-term transmission efficiency.
Under this model, channel allocation may be interpreted as a planning task in a particular class of
Partially Observed Markov Decision Process (POMDP) also called restless
bandits \cite{Liu:Zhao:08,Zhao:al:07bis}.

In the works of \cite{Liu:Zhao:08,Zhao:al:08,Zhao:al:07bis}, it is assumed that the statistical information
about the primary users' traffic is fully available to the secondary user. In practice however, the
statistical characteristics of the traffic 
are not fixed a priori and must be somehow estimated by
the secondary user. 
As the secondary user selects channels to sense, we are not faced with a simple parameter estimation problem
but with a task which is closer to reinforcement learning \cite{Sutton:92}.  
We consider scenarios in which the
secondary user first carries out an \emph{exploration phase} in which the statistical information
regarding the model is gathered and then follows by the \emph{exploitation phase}, where the
optimal sensing policy, based on the estimated parameters, is applied. The key issue is to reach
the proper balance between exploration and exploitation. This issue has been considered before by
\cite{Long:al:08} who proposed an asymptotic rule to set the length of the exploration phase but
without a precise evaluation of the performance of this approach. Lai et al \cite{Lai:al:08} also
considered this problem in the multiple secondary users case but in a simpler model where each
channel is modeled as an independent and identically distributed source. In the field of
reinforcement learning, this class of problems is known as \emph{model-based reinforcement learning}
for which several approaches have been proposed recently
\cite{Auer:Ortner:07,Strehl:Littman:08,Tewari:Bartlett:08}.  
However, none of these
directly applies to the channel allocation model in which the state of the channels is only
partially observed. 

Our contribution consists in proposing a strategy, termed \emph{Tiling Algorithm}, for adaptively
setting the length of the exploration phase. Under this strategy, the length of the exploration
phase is not fixed beforehand and the exploration phase is terminated as soon as we have
accumulated enough statistical evidence to determine the optimal sensing policy. The distinctive
feature of this approach is that it comes with strong performance guarantees in the form of
finite-horizon regret bounds. 
For the sake of clarity, this strategy is described in the general
abstract framework of parametric POMDPs. 
 Remark that the channel access model corresponds to a specific example of POMDP parameterized by the transition probabilities of the availability of each channel.
As the approach relies on the restrictive assumption that
for each possible parameter value the solution of the planning problem be fully known, it is not
applicable to POMDPs at large but is well suited to the case of the channel allocation model. We
provide a detailed account of the use of the approach for two simple instances of the opportunistic channel
access model, including the case of stochastically identical channels considered by
\cite{Zhao:al:08}.

The article is organized as follows. The channel allocation model is formally described in Section
\ref{sec:ChannelModel}. In Section \ref{sec:TA}, the tiling algorithm is presented and its
performance in terms of finite-horizon regret bounds are obtained. The application to opportunistic channel
access is detailed in Section \ref{sec:appl}, both in the one channel model and in the case of
stochastically identical channels.

\section{Channel Access Model}\label{sec:ChannelModel}
Consider a network consisting of $N$ independent channels with time-varying state, with bandwidths
$B(i)$, for $i=1,\dots\,N$. These $N$ channels are licensed to a primary network whose users
communicate according to a synchronous slot structure. At each time slot, channels are either free
or occupied (see Fig.~\ref{fig:channel}).
Consider now a secondary user seeking opportunities of transmitting in the free slots of these $N$
channels without disturbing the primary network.  With limited sensing, a secondary user can only
access a subset of $M\ll N$ channels.  The aim of the secondary user is to leverage this partial
observation of the channels so as to maximize its long-term opportunities of transmission.
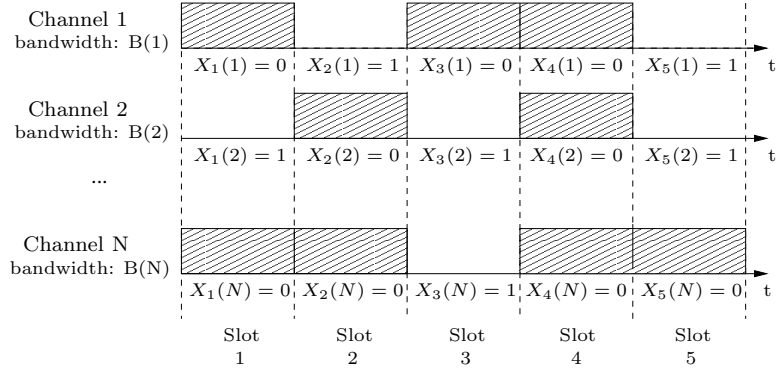
\begin{figure}[h]
  \begin{center}
    \input{channel1ter.pstex_t}
    \caption{Representation of the primary network}
    \label{fig:channel}
  \end{center}
\end{figure}
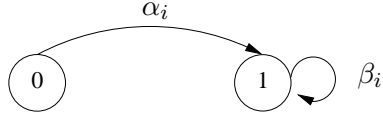
\begin{figure}
  \begin{center}
    \input{transition.pstex_t}
  \end{center}
  \caption{Transition probabilities in the $i$-th channel.}
  \label{fig:transition}
\end{figure}

Introduce the state vector which describes the
network at time $t$, $[X_t(1),\dots,X_t(N)]'$, where $X_t(i)$ is equal to $0$ when the channel $i$
is occupied and $1$ when the channel is idle. The states $X_t(i)$ and $X_t(j)$ of different
channels $i\neq j$ are assumed to be independent. Let $\alpha(i)$ (resp.$\beta(i)$) be the transition probability from state $0$ (resp. $1$) to state $1$ in channel $i$ (see Fig.~\ref{fig:transition}).
Additionally, denote by $(\nu_0(i),\nu_1(i))$ the stationary probability of the Markov chain
$(X_t(i))_t$. The secondary user selects a set of $M$ channels
to sense. This choice corresponds to an action $A_t=[A_t(1),\dots,A_t(N)]'$, where $A_t(i)=1$ if
the $i$-th channel is sensed and $A_t(i)=0$ otherwise. Since only $M$ channels can be sensed,
$\sum_{i=1}^N A_t(i) =M$. The observation
is an $N$-dimensional vector $[Y_t(1),\dots,Y_t(N)]'$ such that $Y_t(i)=X_t(i)$ for the $M$
selected channels and 
$Y_t(i)$ is an arbitrary value not in $\{0,1\}$ for the other channels.  The reward gained at each
time slot is equal to the aggregated bandwidth available. In addition, a reward equal to
$0\leq\lambda\leq\min_iB(i)$ is received for each unobserved channel. At each time $t$, the received reward is
$\sum_{i=1}^N r(X_t(i),A_t(i))$ where
\begin{equation*}
  r(X_t(i),A_t(i))=\begin{cases}
    B(i) &  \text{if $A_t(i)=1$, $X_t(i)=Y_t(i)=1$}\\
    0 &\text{if $A_t(i)=1$, $X_t(i)=Y_t(i)=0$}\\
    \lambda &\text{otherwise}\\
  \end{cases}\eqsp,
\end{equation*}
which depends on $X_t(i)$ only through $Y_t(i)$.
The gain $\lambda$ associated to the action of not observing may also be interpreted as a penalty
for sensing occupied channels.
Indeed, this model is equivalent to the one where a positive reward $B(i)-\lambda$ is received for available sensed channels, a penalty $-\lambda$ is received for occupied sensed channels and no reward are received for non-sensed channels.

Note that this model is a particular POMDP in which the state transition probabilities do not depend on the
actions. Moreover, the independence between the channels may be exploited to construct a
$N$-dimensional \emph{sufficient internal state} which summarizes all past decisions and
observations. The internal state $p_t$ is defined as follows: for all $i\in\{1,\dots N\}$,
$p_t(i)=\CP{X_t(i)=1}{A_{0:t-1}(i), Y_{0:t-1}(i)}$. This internal state enables the secondary user
to select the channels to sense.   The internal state recursion is
\begin{equation}
  p_{t+1}(i)=
  \begin{cases}
    \alpha(i)\qquad \qquad\text{if $A_{t}(i)=1, Y_{t}(i)=0$}\\
    \beta(i) \qquad \qquad\text{if $A_{t}(i)=1, Y_{t}(i)=1$}\\
    p_t(i)\beta(i)+(1-p_t(i))\alpha(i)\quad\text{otherwise}
  \end{cases}\eqsp.
  \label{eq:InternalStateRecursion}
\end{equation}
Moreover, remark that at each time $t$, the internal state $p_t$ is completely defined by the pair
$(k,y)$ where $y=[y(1),\dots,y(N)]'$ denotes the last observed state for each channel and
$k=[k(1),\dots, k(N)]'$ is the duration during which the corresponding channel has not been
observed. Denote by $p^{k(i),y(i)}_{\alpha(i),\beta(i)}$ the probability that a channel is free
given that it has not been observed for $k(i)$ time slots and that the last observation was
$y(i)$. That is to say, for $k(i)>1$,
$ p^{k(i),y(i)}_{\alpha(i),\beta(i)}=\eqsp\P[X_t(i)=1|A_{t-k(i)+1:t-1}(i)=0, A_{t-k(i)}(i)=1,
Y_{t-k(i)}(i)=y(i)]$ and $p^{1,y(i)}_{\alpha(i),\beta(i)}=\CP{X_t(i)=1}{A_{t-1}(i)=1,
  Y_{t-1}(i)=y(i)}\eqsp.$ Using equation \eqref{eq:InternalStateRecursion}, these probabilities may
be written as follows:
\begin{align}
  &p^{k(i),0}_{\alpha(i),\beta(i)}=\frac{\alpha(i)(1-(\beta(i)-\alpha(i))^{k(i)})}{1-\beta(i)+\alpha(i)}\eqsp,\label{eq:pky=0}\\
  &p^{k(i),1}_{\alpha(i),\beta(i)}=\frac{(\beta(i)-\alpha(i))^{k(i)}(1-\beta(i)+\alpha(i))}{1-\beta(i)+\alpha(i)}\eqsp.\label{eq:pky=1}
\end{align}

The channel allocation model may also be interpreted as an instance of the restless multi-armed
bandit framework introduced by \cite{Whittle:88}. Papadimitriou and Tsitsiklis \cite{Papadimitriou:Tsitsiklis:94} have
established that the planning task in the restless bandit model is PSPACE-hard, and hence that
optimal planning is not practically achievable when the number $N$ of channels becomes
important. Nevertheless, recent works have focused on near-optimal so-called \emph{index
  strategies} \cite{LeNy:al:08bis,Guha:Munagala:07,Liu:Zhao:08}, which have a reduced
implementation cost.  An index strategy consists in separating the optimization task into $N$
channel-specific sub-problems, following the idea originally proposed by Whittle \cite{Whittle:88}. 
Interestingly, to determine the Whittle index pertaining to each channel, one
has to solve the planning problem in the single channel model for arbitrary values of
$\lambda$. Using this interpretation, explicit expressions of the Whittle's indexes as a function of the channel transition
probabilities $\{\alpha(i),\beta(i)\}_{i=1,\dots,N}$ have been provided by \cite{LeNy:al:08bis,Liu:Zhao:08}.

\section{The Tiling Algorithm}\label{sec:TA}
Here, we focus on determining the sensing policy when the secondary user does not have any
statistical information about the primary users' traffic. A common approach is to learn the
transition probabilities $\{\alpha(i),\beta(i)\}_{i=1,\dots,N}$ in a first phase and then to act
optimally according to the estimated model.
If the learning phase is sufficiently long, the estimates of the probabilities can be quite precise
and there is a higher chance that the policy followed during the exploitation phase is indeed the
optimal policy. On the other hand, blindly sensing channels to learn the model parameters does not
necessarily coincide with the optimal policy and thus has a cost in terms of performance. The
question is hence: how long should the secondary user learn the model (\emph{explore}) before
applying an \textit{exploitation} policy such as Whittle's policy ?

This problem is the well known dilemma between exploration and exploitation \cite{Sutton:92}. Here we propose an
algorithm to balance exploration and exploitation by adaptively monitoring the duration of the
exploration phase. We present this algorithm in a more abstract framework for generality. We assume
that the optimal policy is a known function of a low dimensional parameter. This condition can be
restrictive but it is verified in simple cases such as finite state space MDPs or in particular
cases of POMDPs like the channel access model (see also Section~\ref{sec:appl}).

\subsection{The Parametric POMDP Model}
Consider a POMDP defined by $(\Xset,\Aset,\Yset,Q_\theta, f, r)$, where $\Xset$ is the discrete state space,
$\Yset$ is the observation space, $\Aset$ is the finite set of actions,
$Q_\theta:\Xset\times\Aset\times\Xset\rightarrow[0,1]$ is the transition probability,
$f:\Xset\times\Aset\rightarrow \Yset$ is the observation function,
$r:\Xset\times\Aset\rightarrow\Rset$ is the bounded reward function and $\theta\in\Theta$ denotes
an unknown parameter. Given the current hidden state $x\in\Xset$ of the system, and a control
action $a\in\Aset$, the probability of the next state $x'\in\Xset$ is given by
$Q_\theta(x,a;x')$. At each time step $t$, one chooses an action $A_t =\pi(A_{0:t-1},Y_{0:t-1})$ according to a \emph{policy} $\pi$, and hence observes $Y_t = f(X_t, A_t)$ and receives the reward
$r(X_t,A_t)$. Without loss of generality, we assume that for all $x\in\Xset$, for all $a\in\Aset$,
$r(x,a)\leq 1$.

Since we are interested in rewards accumulated over finite but large horizons, we will consider the
average (or long-term) reward criterion defined by
\[
V^\pi_\theta =
\lim_{n\rightarrow\infty}\frac{1}{n}\E^\pi_\theta\left(\sum_{t=1}^{n}r(X_t,A_t)\right) \eqsp,
\]
where $\pi$ denotes a fixed policy. The notation $V^\pi_\theta$ is meant to highlight the fact that
the average reward depends on both the policy $\pi$ and the actual parameter value $\theta$. For a
given parameter value, the optimal long-term reward is defined as $ V^*_\theta=\sup_\pi
V^\pi_\theta $ and $\pi^*_\theta$ denotes the associated optimal policy. We assume that the
dependence of $V^\pi_\theta$ and $\pi^*_\theta$ with respect to $\theta$ is fully known. In
addition, there exists a particular default policy $\pi_0$ under which the parameter $\theta$ can
be consistently estimated.

Given the above, one can partition the parameter space $\Theta$ into non-intersecting subsets,
$\Theta=\bigcup_i Z_i$, such that each policy zone $Z_i$ corresponds to a single optimal policy,
which we denote by $\pi_i^*$. In other words, for any $\theta\in Z_i$, $
V^*_\theta=V^{\pi_i^*}_\theta.  $ In each policy zone $Z_i$, the corresponding optimal policy
$\pi^*_i$ is assumed to be known as well as the long-term reward function $V^{\pi^*_i}_\theta$ for
any $\theta\in\Theta$.

\subsection{The Tiling Algorithm (TA)}
We denote by $\hat\theta_t$ the parameter estimate obtained after $t$ steps of the
exploration policy and by $\Delta_t$ the associated confidence region, whose construction will be
made more precise below.  The principle of the tiling algorithm is to use the policy zones
$(Z_i)_i$ to determine the length of the exploration phase: basically, the exploration phase will
last until the estimated confidence region $\Delta_t$ fully enters one of the policy zones. It
turns out however that this naive principle does not allow for a sufficient control of the expected
duration of the exploration phase, and, hence, of the algorithm's regret. In order to deal with
parameter values located close to the borders of policy zones, one needs to introduce additional
\emph{frontier zones} $(F_j(n))_j$ that will shrink at a suitable rate with the time horizon
$n$. 
Let
\begin{equation}
  T_n=\inf\{t \geq 1 : \exists i,\eqsp \Delta_t\subset Z_i\eqsp \text{or }\exists j,\eqsp
  \Delta_t\subset F_j(n)\}
  \label{eq:end_explore}
\end{equation}
denote the random instant where the exploration terminates. Note that the frontier zones
$(F_j(n))_j$ depends on $n$. Indeed, the larger $n$ the smaller the frontier zones can be in order
to balance the length of the exploration phase and the loss due to the possible choice of a
suboptimal policy.
\begin{figure}[hbtp] \centering
  \includegraphics[width=0.42\textwidth]{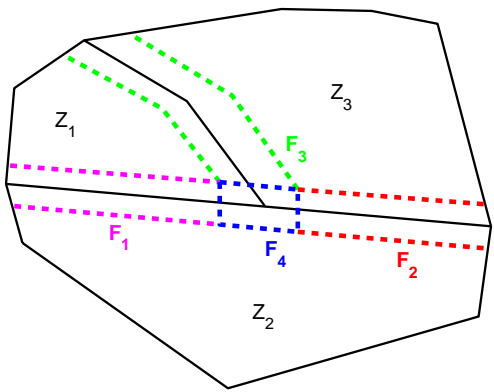}
  \caption{Tiling of the parameter space for an example with three distinct optimal policy zones.}
  \label{fig:zones_and_frontiers}
\end{figure}

In Figure \ref{fig:zones_and_frontiers}, we represent the tiling of the parameter space for an
hypothetical example with three distinct optimal policy zones. In this case, there are four
frontier zones: one between each pair of policy zones ($F_1(n)$, $F_2(n)$ and $F_3(n)$) and another
($F_4(n)$) for the intersection of all the policy zones. In the following, we shall assume that
there exists only finitely many distinct frontier and policy zones.

The tiling algorithm consists in using the default exploratory policy $\pi_0$ until the occurrence
of the stopping time $T_n$, according to~\eqref{eq:end_explore}. From $T_n$ onward, the algorithm
then selects 
a policy to use during the remaining time as follows:
if at the end of the exploration phase, the confidence region is fully included in a policy zone $Z_i$,
then the selected policy is $\pi^*_i$;
otherwise, the confidence region is included in a frontier zone $ F_j(n)$ and the selected
policy is any optimal policy $\pi^*_k$ \emph{compatible} with the frontier zone
$F_j(n)$.
An optimal policy $\pi^*_k$ is said to be \emph{compatible} with the frontier zone $F_j(n)$ if the
intersection between the policy zone $Z_k$ and the frontier zone is non empty.
In the example of Figure~\ref{fig:zones_and_frontiers}, for instance, $\pi^*_1$ and $\pi^*_2$ are
compatible with the frontier zone $F_1(n)$, while all the optimal policies $(\pi^*_i)_{i=1,2,3}$
are compatible with the central frontier zone $F_4(n)$. If the exploration terminates in a frontier
zone, then one basically does not have enough statistical evidence to favor a particular optimal
policy and the tiling algorithm simply selects one of the optimal policies compatible with the
frontier zone. Hence, the purpose of frontier zones is to guarantee that the exploration phase will
stop even for parameter values for which discriminating between several neighboring optimal
policies is challenging. Of course, in practice, there may be other considerations that suggest to
select one compatible policy rather than another but the general regret bound below simply assumes
that any compatible policy is selected at the termination of the exploration phase.

\subsection{Performance Analysis}
To evaluate the performance of this algorithm, we will consider the regret, for the prescribed time
horizon $n$, defined as the difference between the expected cumulated reward obtained under the
optimal policy and the one obtained following the algorithm,
\begin{equation}
  R_n(\theta^*)=\E^{\pi^*_{\theta^*}}_{\theta^*}\left[\sum_{t=1}^n r(X_t,A_t)\right] - \E^{\mathrm{TA}}_{\theta^*} \left[\sum_{t=1}^n r(X_t,A_t)\right] \eqsp, 
  \label{eq:regret}
\end{equation}
where $\theta^*$ denotes the unknown parameter value. To obtain bounds for $R_n(\theta^*)$ that do
not depend on $\theta^*$, we will need the following assumptions.
\begin{assumption}
  \label{assum:confidence}
  The confidence region $\Delta_t$ is constructed so that there exists constants $c_1,c_1',
  n_{\min}\in\Rset_+$ such that, for all $\theta\in\Theta$, for all $n\geq n_{\min}$, for all
  $t\leq n$,
  $ \P_{\theta}\left(\theta \in \Delta_t,\eqsp \delta(\Delta_t)\leq c_1 \frac{\sqrt{\log
        n}}{\sqrt{t}}\right)\geq 1-c'_1 \exp\{-\frac{1}{3}\log n\}\eqsp, $ where
  $\delta(\Delta_t)=\sup\{\normi{\theta-\theta'},\eqsp\theta,\theta'\in \Delta_t\}$ is the diameter of the confidence region.
\end{assumption}

\begin{assumption}
  \label{assum:zones}
  Given a size $\epsilon(n)$, one may construct the frontier zones $(F_j(n))_j$ such that there
  exists constants $c_2,c_2'\in\Rset_+$ for which
  \begin{itemize}
  \item $\delta(\Delta_t)\leq c_2 \epsilon(n)$ implies that there exists either $i$ such that
    $\Delta_t\subset Z_i$ or $j$ such that $\Delta_t\subset F_j(n)$,
  \item if $\theta\in F_j(n)$, there exists $\theta'\in Z_i$ such that $\normi{\theta-\theta'} \leq
    c_2'\epsilon(n)$, for all policy zones $Z_i$ compatible with $F_j(n)$ (i.e., such that $Z_i
    \bigcap F_j(n)\neq\emptyset$).
  \end{itemize}
\end{assumption}
\begin{assumption}
  \label{assum:regularity}
  For all $i$, there exists $d_i \in \Rset_+$ such that for all $\theta, \theta' \in \Theta$, $
  |V_\theta^{\pi_i^*}-V_{\theta'}^{\pi_i^*}|\leq d_i\normi{\theta-\theta'}\eqsp.  $
\end{assumption}

Assumption~\ref{assum:confidence} pertains to the construction of the confidence region and may
usually be met by standard applications of the Hoeffding inequality. The constant $1/3$ is meant to
match the worst-case rate given in Theorem~\ref{theo:main} below. Assumption~\ref{assum:zones}
formalizes the idea that the frontier zones should allow any confidence region of diameter less
than $\epsilon(n)$ to be fully included either in an original policy zone or in a frontier zone,
while at the same time ensuring that, locally, the size of the frontier is of order
$\epsilon(n)$. The applicability of the tiling algorithm crucially depends on the construction of
these frontiers.  Finally, Assumption~\ref{assum:regularity} is a standard regularity condition
(Lipschitz continuity) which is usually met in most applications. 
The performance of the tiling approach is given by the following theorem, which
is proved in Appendix~\ref{appendix:proofTheorem}.
\begin{theorem}
  \label{theo:main}
  Under assumptions~\ref{assum:confidence}, \ref{assum:zones} and~\ref{assum:regularity}, and for
  all $n\geq n_{\min}$, the duration of the exploration phase is bounded, in expectation, by
  \begin{equation}
    \E_{\theta^*}(T_n)\leq c\frac{\log n}{\epsilon^2(n)} \eqsp ,
    \label{eq:duration_bound}    
  \end{equation}
  and the regret by
  \begin{equation}
    R_n(\theta^*)\leq \E_{\theta^*}(T_n) + c'n \epsilon(n) + c''n \exp\{-\frac{1}{3}\log n\} \eqsp,
    \label{eq:regret_bound}   
  \end{equation}
  where $c=(c_1/c_2)^2$, $c'=c'_2\max_{i,k}(d_i+d_k)$ and $c''=c'_1$.
  The minimal worst-case regret is obtained when selecting $\epsilon(n)$ of the order of $(\log n
  /n)^{1/3}$, which yields the bound $
  R_n(\theta^*) \leq C (\log n)^{1/3} \, n^{2/3} $ 
  for some constant $C$.
\end{theorem}

The duration bound in~\eqref{eq:duration_bound} follows from the observation that exploration is
guaranteed to terminate only when the confidence region defined by
Assumption~\ref{assum:confidence} reaches a size which is of the order of the diameter of the
frontier, that is, $\epsilon(n)$. The second term in the right-hand side of~\eqref{eq:regret_bound}
corresponds to the maximal regret if the exploration terminates in a frontier zone. The rate $(\log
n)^{1/3}n^{2/3}$ is obtained when balancing these two terms ($\E_{\theta^*}(T_n)$ and $c'n
\epsilon(n)$). A closer examination of the proof in Appendix~\ref{appendix:proofTheorem} shows that
if one can ensure that the exploration indeed terminates in one of the policy regions $Z_i$, then
the regret may be bounded by an expression similar to~\eqref{eq:regret_bound} but without the $ c'n
\epsilon(n)$ term. In this case, by using a constant strictly larger than 1---instead of $1/3$---in
Assumption~\ref{assum:confidence}, one can obtain logarithmic regret bounds. To do so, one however
need to introduce additional constraints to guarantee that exploration terminates into a policy
region rather than in a frontier. These constraints typically take the form of an assumed
sufficient margin between the actual parameter value $\theta^*$ and the borders of the associated
policy zone. This is formalized in Theorem~\ref{theo:RegretLog} which is proved in
Appendix~\ref{appendix:proofTheoRegretLog}. First, introduce an alternative of\note{SF}{à rédiger
  mieux} Assumption~\ref{assum:confidence}.

\begin{assumption}\label{assum:confidence_bis}
  The confidence region $\Delta_t$ is constructed so that there exists constants
  $c_1,c_1',n_{\min}\in\Rset_+$, $x>1$ such that, for all $\theta\in\Theta$, for all $n\geq
  n_{\min}$, for all $t\leq n$,
  $\P_{\theta}\left(\theta \in \Delta_t,\eqsp \delta(\Delta_t)\leq c_1
    \frac{\sqrt{x}}{\sqrt{t}}\right)\geq 1-c'_1 \exp\{-2x\}\eqsp.$
\end{assumption}
\begin{theorem}
  \label{theo:RegretLog}
  Consider $\theta^*$ in a policy zone $Z$ such that there exists $\kappa$ for which
  $\min_{\theta\notin Z}\normi{\theta^*-\theta}> \kappa$. Under assumption~\ref{assum:confidence_bis}, the
  regret is bounded by $R_n(\theta^*)\leq C(\kappa) \log(n)+C'(\kappa)$ for all
  $n\geq n_{\min}$ and for some constants $C(\kappa)$ and $C'(\kappa)$ which decrease with $\kappa$.
\end{theorem}

\section{Application to Channel Access}
\label{sec:appl}

In the following,
we consider two specific instances of the opportunistic channel access model introduced in
Section~\ref{sec:ChannelModel}. First, we study the single channel case which is an interesting
illustration of the tiling algorithm. Indeed, in this model, there are a lot of different policy
zones and both the optimal policy and the long-term reward can be explicitly computed in each of
them. In addition, the one channel model plays a crucial role in determining the Whittle index
policy. Next, we apply the tiling algorithm to a $N$ channel model with stochastically
identical channels.

\subsection{One Channel Model}\label{sec:1channel}
Consider a single channel with bandwidth
$B=1$. At each time, the secondary user can choose to sense the channel hoping to receive a reward equal
to $1$ if the channel is idle and taking the risk of receiving no reward if the channel is
occupied. He can also decide to not observe the channel and then to receive a reward equal to
$0\leq\lambda\leq 1$.

\subsubsection{Optimal policies, long-term rewards and policy zones}
\begin{figure}[hbtp]
  \centering
  \includegraphics[width=0.5\textwidth]{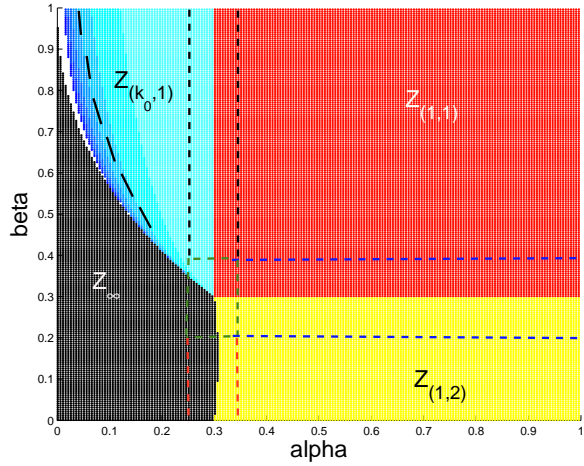}
  \caption{The optimal policy regions in the one channel model with $\lambda=0.3$.}
  \label{fig:PoliciesregionAlphaBeta}
\end{figure}

Studying the form of the optimal policy as a function of $\theta = (\alpha,\beta)$ brings to light
several optimal policy zones.\note{SF}{Mettre une référence ?} In each zone, the optimal policy is
different and is characterized by the pair $(k_0,k_1)$ which defines how long the secondary user
needs to wait (i.e. not observe the channel) before observing the
channel again depending on the outcome of the last observation. Denote by $\pi^*_{(k_0,k_1)}$ the policy which consists in waiting $k_0-1$
(resp. $k_1-1$) time slots before observing the channel again if, last time the channel was sensed, it
was occupied (resp. idle), and by $Z_{(k_0,k_1)}$ the corresponding policy zone. Let $\pi^*_\infty$
be the policy which consists in never observing the channel; this policy is optimal when $\alpha$ and $\beta$ are such that the probability that the channel is idle is always lower than $\lambda$. We represent in Figure
\ref{fig:PoliciesregionAlphaBeta} the policy zones.

The long-term reward of each policy can be exactly computed:
\begin{align*}
  &V^{\pi_{(1,1)}^*}_{\alpha,\beta}=\frac{\alpha}{1-\beta+\alpha}\eqsp,\\
  &V^{\pi_{(1,2)}^*}_{\alpha,\beta}=\alpha\frac{1+\lambda}{1+\alpha+\beta(\alpha-\beta)}\eqsp,\\
  &V^{\pi_{(k_0,1)}^*}_{\alpha,\beta}=\frac{(k_0-1)(1-\beta)\lambda+1\,p^{k_0,0}_{\alpha,\beta}}{k_0(1-\beta)+p^{k_0,0}_{\alpha,\beta}}\eqsp,\text{ for $k_0\geq 2$,}\\
  &V^{\pi_{\infty}^*}_{\alpha,\beta}=\lambda\eqsp.\\
\end{align*}

\subsubsection{Applying the tiling algorithm}
Applying the tiling algorithm to this model is not straightforward as there are an infinity of
policy zones. We introduce border
zones between $Z_{(1,1)}$, $Z_{(2,1)}$, $Z_{(1,2)}$, $Z_\infty$ as shown in
Figure~\ref{fig:PoliciesregionAlphaBeta}. Moreover, to address the problem of the infinity of zones, we propose to aggregate the policy zones when
$\alpha<\lambda$ and $\beta>\lambda$. For example, we aggregate all the zones $Z_{(k_0,1)}$ with
$2\leq k_0\leq l$ and the non-observation zone $Z_\infty$ with the zones $Z_{(k_0,1)}$ such that
$k_0\geq l'$, where $l'\leq l$ are variables to be tuned according to the time horizon $n$. Thus, Theorem~\ref{theo:main} still applies.

Recall that the tiling algorithm consists in learning the
parameter $(\alpha,\beta)$ until the estimated confidence region fully enters either one of the
policy zones or one of the frontier zones. The exploration policy, denoted by $\pi_0$ in
Section~\ref{sec:TA}, consists in always sensing the channel. At time $t$, the estimated parameter is given by
\begin{equation}
  \hat\alpha_t = \frac{N_t^{0,1}}{N_t^0}\eqsp \text{ and }\eqsp \hat\beta_t = \frac{N_t^{1,1}}{N_t^1}\eqsp,\label{eq:estimators}
\end{equation}
where $N_t^0$ (resp. $N_t^1$) is the number of visits to $0$ (resp. $1$) until time $t$ and
$N_t^{0,1}$ (resp. $N_t^{1,1}$) is the number of visits to $0$ (resp. $1$) followed by a visit to
$1$ until time $t$.

In order to verify that this model satisfies the conditions of Theorem~\ref{theo:main}, we need to
make an irreducibility assumption on the Markov chain.
\begin{assumption}
  \label{assum:theta}
  There exists $\eta$ such that $(\alpha,\beta)\in\Theta=[\eta,1-\eta]^2$.
\end{assumption}
This condition ensures that, during the time horizon $n$, the Markov chain visits the two states
sufficiently often to estimate the parameter $(\alpha,\beta)$. We define the confidence region as the
rectangle
\begin{equation}
  \Delta_t=\left[\hat\alpha_t\pm \sqrt{\frac{\log n}{6 N_t^0}}\right]\times\left[\hat\beta_t\pm \sqrt{\frac{\log n}{6 N_t^1}}\right]\eqsp.
  \label{eq:delta}
\end{equation}

To prove that the regret of the tiling algorithm in a single channel model is bounded, we need to
verify the three assumptions of Theorem~\ref{theo:main}. First, it is shown in
appendix~\ref{appendix:confidenceIntervalMC} that Assumption~\ref{assum:confidence} holds.
Secondly, except when $\alpha<\lambda$ and $\beta>\lambda$, Assumption~\ref{assum:zones} is
obviously satisfied, since the confidence region and the policy and frontier zones are all
rectangles (see Fig.~\ref{fig:PoliciesregionAlphaBeta}). Let $\epsilon(n)$ be half of the smallest width of the frontier zones. Additionally,
when $\alpha<\lambda$ and $\beta>\lambda$, if the center frontier zone is large enough, the
aggregation of the zones can be done such that the second condition holds. Finally, for all optimal
policy, the long-term reward is a Lipschitz continuous function of $(\alpha, \beta)$ for $\alpha,
\beta \in [\eta,1-\eta]$, so the third condition is also satisfied.

\subsubsection{Experimental results}
\begin{figure}[hbtp]
  \begin{center}
    \includegraphics[width=0.5\columnwidth]{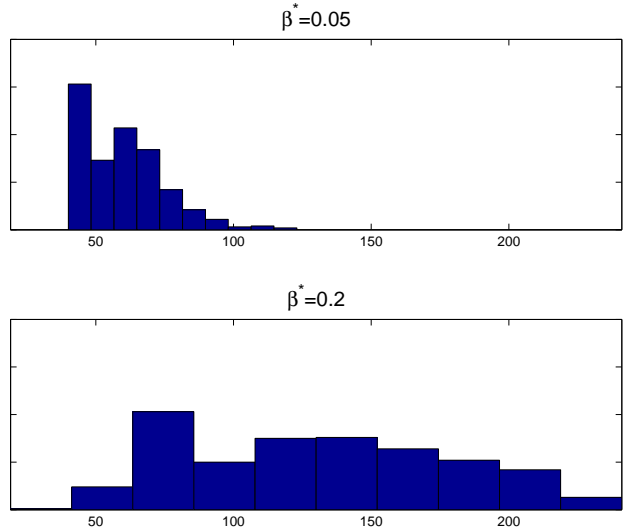}
    \caption{Distribution of the length of the exploration phase following the tiling algorithm for
      $(\alpha^*,\beta^*)=(0.8, 0.05)$ and for $(\alpha^*,\beta^*)=(0.8, 0.2)$.}
    \label{fig:explLengthTA1channel}
  \end{center}
\end{figure}

As suggested by Theorems~\ref{theo:main}--~\ref{theo:RegretLog}, the length of the exploration
phase following the tiling algorithm depends on the value of the true parameter
$(\alpha^*,\beta^*)$. 
In addition, for a fixed value of $(\alpha^*,\beta^*)$, the length of the exploration varies from
one run to another, depending on the size of the confidence region. To illustrate these effects, we
consider two different value of the parameters: $(\alpha^*,\beta^*)=(0.8, 0.05)$ which is included
in the policy zone $Z_{(1,2)}$ and far from any frontier zone, and, $(\alpha^*,\beta^*)=(0.8, 0.2)$
which lies in the frontier zone between $Z_{(1,1)}$ and $Z_{(1,2)}$ and is close to the border of
the frontier zone. The corresponding empirical distributions of the length of the exploration phase
are represented in Figure~\ref{fig:explLengthTA1channel}. Remark that 
the shape of these two distributions are quite different and that the empirical mean of the length of the exploration phase is lower for a parameter which is far from any frontier zone than for a parameter which is close to the border of a frontier zone.

\begin{figure}[hbtp]
  \begin{center}
    \includegraphics[width=0.5\columnwidth]{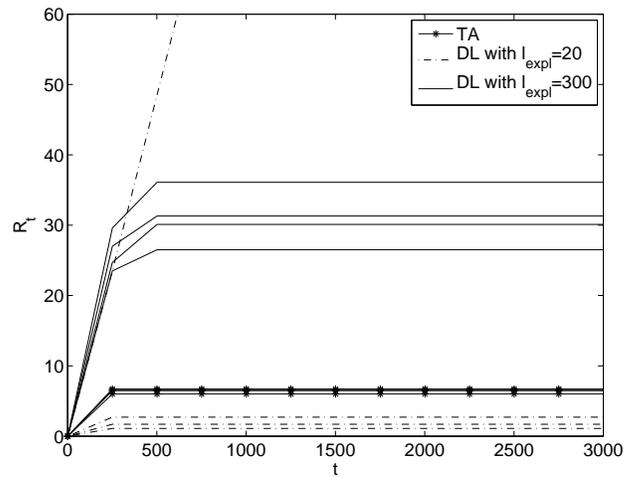}
    \caption{Comparison of the cumulated regret of the tiling algorithm (shaped markers) and an
      algorithm with a deterministic length of exploration phase equal to 20 (dashed line) or equal
      to 300 (solid line) for $(\alpha^*,\beta^*)=(0.8, 0.05)$}
    \label{fig:regret_compareExplDet}
  \end{center}
\end{figure}

In Figure~\ref{fig:regret_compareExplDet}, we compare the cumulated regrets $R_n^{TA}$ of the tiling algorithm to the regrets $R_n^{DL}(l_{expl})$ of an algorithm
with a deterministic length of exploration phase $l_{expl}$. Both algorithms are run with
$(\alpha^*,\beta^*)=(0.8,0.05)$. We use two values of $l_{expl}$: one lower ($l_{expl}=20$) and the
other larger ($l_{expl}=300$) than the average length of the exploration phase following the tiling
algorithm which ranges between $40$ and $150$ for this value of the parameter (see Fig.~\ref{fig:explLengthTA1channel}). The algorithms are run four times independently and every cumulated regret are represented in Figure~\ref{fig:regret_compareExplDet}.
Note that, $(\alpha^*,\beta^*)$ being in the interior of a policy zone (i.e. not in a frontier
zone), the regret of the tiling algorithm is null during the exploitation phase since the optimal policy for the true
parameter is used. 
Similarly, when the deterministic length $l_{expl}$ of the exploration phase is sufficiently large, the estimation of the parameter is quite precise, therefore the regret during the exploitation phase is null. On the other hand, too large a value of $l_{expl}$ increases the regret during the exploration phase: we oberve in Figure~~\ref{fig:regret_compareExplDet} that the regret $R_n^{DL}(l_{expl})$ with $l_{expl}=300$ is larger than $R_n^{TA}$. 
When the deterministic length of the exploration phase is smaller than the average length of the exploration phase following the tiling
algorithm, either the parameter is estimated precisely enough and then $R_n^{DL}(l_{expl})$ is smaller than $R_n^{TA}$, or, the estimated  value is too far away from the actual value and the policy followed during the exploitation phase is not the optimal one. In the latter case, the regret is not null during the exploitation phase and $R_n^{DL}(l_{expl})$ is noticeably large. This can be observed in Figure~~\ref{fig:regret_compareExplDet}: in three of the
four runs, the cumulated regret $R_n^DL(l_{expl})$ with $l_{expl}=20$ (dashed line) are small, whereas in the remaining run it sharply and constantly increases.

\subsection{Stochastically Identical Channels Case}
In this section, consider a full channel allocation model where all the $N$ channels have equal
bandwidth $B=1$ and are stochastically identical in terms of primary usage, i.e. all the channels
have the same transition probabilities: $\forall i\in\{1,\dots,
N\}\eqsp,\:\alpha_i=\alpha\eqsp,\:\beta_i=\beta\eqsp.$ In addition, let $\lambda=0$.

\subsubsection{Optimal policies, long-term rewards and policy zones}
Under these assumptions, the near optimal Whittle's index policy has been shown to be
equivalent to the \emph{myopic policy} (see \cite{Liu:Zhao:08}) which consists in selecting the channels to be sensed
according to the expected one-step reward: $A_t= \argmax_{a\in\Aset}\sum_{i=1}^N
a(i)p^{k(i),y(i)}_{\alpha,\beta}\eqsp$ given that channel $i$ has not been observed for $k(i)$ time slots and the last observation was $y(i)$. Recall that $\Aset$ denote the set of
$N$-dimensional vectors with $M$ components equal to $1$ and $N-M$ equal to $0$.  Following this policy, the secondary user senses the $M$
channels that have the highest probabilities $p_{\alpha,\beta}^{k(i),y(i)}$ to be free.

\begin{figure}[hbtp] \centering
  \includegraphics[width=0.42\textwidth]{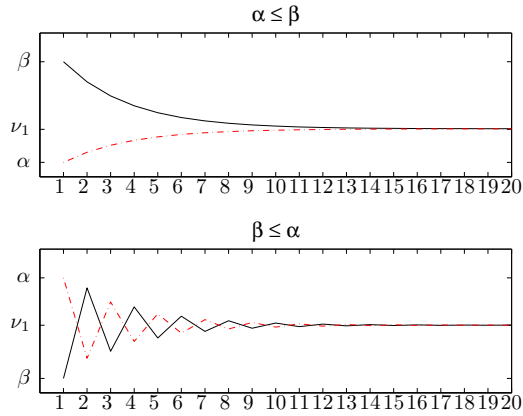}
  \caption{Probabilities $p^{k(j),y(j)}_{\alpha,\beta}$ that the $j$-th channel is idle for
    $y(j)=1$ (solid line) and $y(j)=0$ (dashed line) as a function of $k(j)$, in the positively
    (top) and the negatively (bottom) correlated cases.}
  \label{fig:internal}
\end{figure}

The resulting policy depends only on whether the system is positively correlated ($\alpha\leq
\beta$) or negatively correlated ($\beta\leq\alpha$) (see \cite{Liu:Zhao:08} for details). To explain an important difference
between the positively and negatively correlated cases, we represent in Figure~\ref{fig:internal} the
probability $p^{k(j),y(j)}_{\alpha,\beta}$ that the $j$-th channel is idle for $y(j)=1$ and $y(j)=0$ as a function of $k(j)$, in the two cases.  We observe that, for
all $k(j)\geq 1$, for all $y(j)\in\{0,1\}$,
\begin{equation}
  \begin{cases}
    p^{1,0}_{\alpha,\beta}=\alpha\leq p^{k(j),y(j)}_{\alpha,\beta}\leq\beta=p^{1,1}_{\alpha,\beta}&\text{if $\alpha\leq\beta$}\eqsp,\label{eq:ProbaStochId}\\
    p^{1,1}_{\alpha,\beta}=\beta\leq
    p^{k(j),y(j)}_{\alpha,\beta}\leq\alpha=p^{1,0}_{\alpha,\beta}&\text{if $\beta\leq\alpha$}\eqsp.
  \end{cases}
\end{equation}
Then, in the positively correlated case, according to equation~\eqref{eq:ProbaStochId}, if a
channel $i$ has just been observed to be idle, i.e. $k(i)=1,\eqsp y(i)=1$, the optimal action is to
observe it once more since the channel has the highest (or equal) probability to be free: for all
$j\neq i$, $p^{k(i),y(i)}_{\alpha,\beta}\geq p^{k(j),y(j)}_{\alpha,\beta}$. On the contrary, if a
channel has just been observed to be occupied, i.e. $k(i)=1,\eqsp y(i)=0$, it is optimal to not
observe it since the channel has the lowest probability to be free.  When the system is negatively
correlated, the policy is reversed.

Let $\pi_+$ be the policy in the positively correlated case and $\pi_-$ the policy in the
negatively correlated one.
The long-term reward of policies $\pi_+$ and $\pi_-$ can not be computed exactly. However, one may
use the approach of \cite{Zhao:al:08} to compute an approximation of $V_{\alpha,\beta}^{\pi_+}$ and
$V_{\alpha,\beta}^{\pi_-}$ and obtain:
\begin{equation}
  \begin{cases}
    &V_{\alpha,\beta}^{\pi_+}\approx M\frac{\nu_1}{1-\beta+\nu_1}\eqsp,\label{eq:Vapprox}\\
    &V_{\alpha,\beta}^{\pi_-}\approx M\frac{\alpha}{1-\nu_1+\alpha} \eqsp.
  \end{cases}
\end{equation}

\subsubsection{Applying the tiling algorithm}
\begin{figure}[hbtp] \centering
  \includegraphics[width=0.42\textwidth]{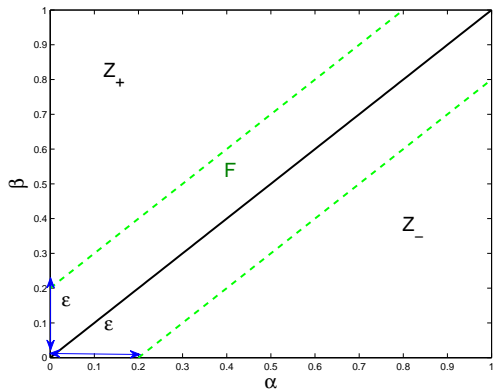}
  \caption{Policy zones and frontier for the N stochastically identical channels model.}
  \label{fig:zones_Nchannels}
\end{figure}

The secondary user thus needs to distinguish between values of the
parameter that lead to positive or negative one-step correlations in the chain. Knowing which of
these two alternatives applies is sufficient to determine the optimal policy. Let $Z_+$ and $Z_-$
be the policy zones corresponding to these two optimal policies $\pi_+$ and $\pi_-$ (see
Figure~\ref{fig:zones_Nchannels}). Between these zones, we introduce a frontier zone
$F(n)=\{(\alpha,\beta), |\alpha-\beta|\leq\epsilon(n)\}$.

The estimation of the parameter $(\alpha,\beta)$ and the confidence region are similar to the one
channel case (see Section~\ref{sec:1channel}). The Assumption~\ref{assum:confidence} of
Theorem~\ref{theo:main} is thus satisfied. Moreover, given the simple geometry of the frontier zone,
Assumption~\ref{assum:zones} is easily verified.  Indeed, any confidence rectangle whose length
is less than $\epsilon(n)/2$ is either included in the frontier zone or in one of the
policy zones. Moreover, for any point in the frontier zone, there exists a point which is at a
distance less than $\epsilon(n)$ and is also in the frontier zone but belongs to the other policy
zone.  Finally, the approximations of the long-term rewards
$V^{\pi_+}_{\alpha,\beta}$ and $V^{\pi_-}_{\alpha,\beta}$ defined in~\eqref{eq:Vapprox} are
Lipschitz functions, and hence the third condition of Theorem~\ref{theo:main} is satisfied.

\subsubsection{Experimental Results}

\begin{figure}[hbtp] \centering
  \includegraphics[width=0.5\textwidth]{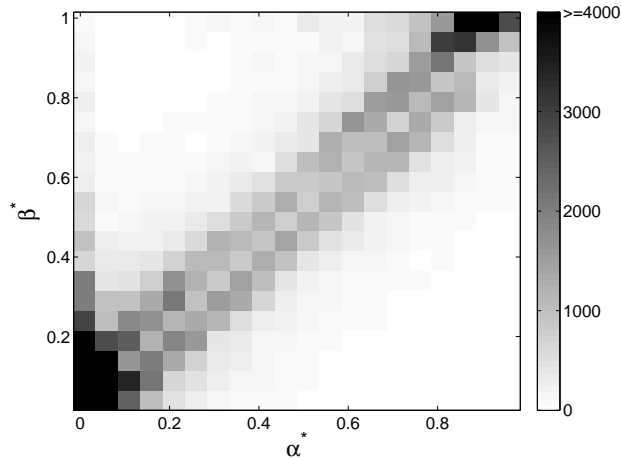}
  \caption{ Length of the exploration phase for the tiling algorithm for different values of
    $(\alpha^*,\beta^*)$.}
  \label{fig:Duree_expl_T10000}
\end{figure}

To illustrate the performance of the approach, we ran the tiling algorithm for a grid of values of
$(\alpha^*,\beta^*)$ regularly covering the set $[\eta,1-\eta]$, with $\eta = 0.01$. For each value
of the parameter, 10 Monte Carlo replications of the data were processed. The time horizon is
$n=10,000$ and the width $\epsilon(n)$ of the frontier zone is taken equal to 0.15. The resulting
cumulated regret has an empirical distribution which does not vary much with the actual value of the
parameter and is, on average, smaller than $90$. However, it may be observed that the average
length of the exploration phase $T_n$, represented in Figure~\ref{fig:Duree_expl_T10000}, depends
on the value of $(\alpha^*,\beta^*)$. First observe that $T_n$ is quite large for
$(\alpha^*,\beta^*)$ close to the frontier zone and small otherwise. Indeed, when the actual
parameter is far from the policy frontier, the exploration phase runs until the confidence region
is included in the corresponding policy zone, which is achieved very rapidly. On the contrary, when
the true parameter is inside the frontier zone, the exploration phase lasts longer. Remark that for
parameter values that sit exactly on the policy frontier both policies are indeed equivalent. This
observation is captured, to some extent, by the algorithm as the maximal durations of the
exploration phase do not occur exactly on the policy frontier. The second important observation is
that the exploration phase is the longest when $(\alpha^*,\beta^*)$ is close to $(0,0)$ or $(1,1)$.
Actually, when $(\alpha^*,\beta^*)$ is around $(0,0)$ (resp. $(1,1)$), the channel is really often
busy (resp. idle) and hence it is difficult to estimate $\beta$ (resp. $\alpha$).

The later effect is partially predicted by the asymptotic approach of \cite{Long:al:08} who used
the Central Limit Theorem to show that the length of the exploration phase, for a channel with
transition probabilities $(\alpha^*,\beta^*)$, has to be equal to
\begin{equation}
  l_{expl}(\alpha^*,\beta^*,\delta,P_C) = \frac{(\Phi^{-1}(\frac{P_C+1}{2}))^2}{\delta^2}(1-\alpha^*)(\frac{1}{\alpha^*}+\frac{1}{1-\beta^*}) \label{eq:LongAl}
\end{equation}
in order to guarantee that $\alpha$ is properly estimated (with a similar result holding for
$\beta$). In~\eqref{eq:LongAl} $\Phi$ stands for the standard normal cumulative distribution
function and $\delta$ and $P_C$ are values such that
$P_C=\P(|\hat\alpha-\alpha^*|<\delta\alpha^*)$. This formula rightly suggests that when $\alpha^*$
is very small, there are very few observed transitions from the busy to the idle state and hence
that estimating $\alpha$ is a difficult task. However, it can be seen on
Figure~\ref{fig:Duree_expl_T10000} that with the tiling algorithm, the length of the exploration
phase is actually longer when \emph{both} $\alpha$ and $\beta$ are very small but is not
particularly long when $\alpha$ is small and $\beta$ is close to one (upper left corner in
Figure~\ref{fig:Duree_expl_T10000}). Indeed in the latter case, the channel state is very
persistent, which imply few observed transitions and, correlatively, that estimating either
$\alpha$ or $\beta$ would necessitate many observation. On the other hand, in this case the channel
is strongly positively correlated and even a few observations suffice to decide that the
appropriate policy is $\pi_+$ rather than $\pi_-$.
 
\section{Conclusion}
The tiling algorithm is a model-based reinforcement learning algorithm applicable to opportunistic
channel access. This algorithm is meant to adequately balance exploration and exploitation by
adaptively monitoring the duration of the exploration phase so as to guarantee a $(\log n)^{1/3} \,
n^{2/3}$ worst-case regret bounds for a pre-specified finite horizon $n$. Furthermore, it has been
shown in Theorem~\ref{theo:RegretLog} that in large regions of the parameter space, the regret can
indeed be guaranteed to be logarithmic. In numerical experiments on the single channel and
stochastically identical channels models, it has been observed that the tiling algorithm is indeed
able to adapt the length of the exploration phase, depending on the sequence of
observations. Furthermore, we observed in the stochastically identical model that the algorithm was
able to interrupt the exploration phase rapidly in cases where the nature of the optimal policy is
rather obvious.

For the future, the tiling algorithm  promises as well a high potential for other applications for example in wireless communications. 
Concerning the opportunistic channel access, the algorithm as it stands is not able to handle the general $N$ channel model presented Section~\ref{sec:ChannelModel} (with stochastically non-identical channels). 
However, another interesting prospective work would be to adapt our approach such that its main principles apply to the general model.

\appendix
\section{Appendix: Proof of Theorem~\ref{theo:main}}
\label{appendix:proofTheorem}
The confidence zone is such that, at the end of the exploration phase,
$\P_{\theta^*}\left(\theta^*\in \Delta_t\eqsp,\eqsp \delta(\Delta_t)\leq c_1 \sqrt{\log n}/\sqrt{t}\right)\geq 1- c'_1 \exp\{-\frac{1}{3}\log n\}\eqsp.$
At the end of the exploration phase, if the true parameter $\theta^*$ is in the confidence region,
there are two possibilities: either the confidence zone $\Delta_t$ is included in a policy zone
$Z_i$ or it is included in a frontier zone $F_j(n)$. If the confidence zone is in a policy region,
the regret is equal to the sum of the duration of the exploration phase and of the loss
corresponding to the case where the confidence region is violated: $
R_n(\theta^*)=\E_{\theta^*}(T_n)+c_1'n\exp\{-\frac{1}{3}\log n\}\eqsp.  $ If the confidence zone is
in a frontier region $F_j(n)$, an additional term of the regret is the loss due to the fact that
the policy selected at the end of the exploration phase is not necessarily the optimal one for the
true parameter $\theta^*$. Let $\pi_i^*$ denote the optimal policy for $\theta^*$ and $\pi^*_k$
the selected policy. Note that $Z_i$ and $Z_k$ are compatible with $F_j(n)$. The loss is
$V_{\theta^*}^{\pi_i^*}-V_{\theta^*}^{\pi_k^*}=(V_{\theta^*}^{\pi_i^*}-V_{\theta}^{\pi_i^*})+(V_{\theta}^{\pi_k^*}-V_{\theta^*}^{\pi_k^*})+(V_{\theta}^{\pi_i^*}-V_{\theta}^{\pi_k^*})\eqsp,$
where $\theta\in Z_k\bigcap F_j(n)$.  The last term is negative since $\pi^*_k$ is the optimal
policy for $\theta$. The two other terms can be bounded using
Assumption~\ref{assum:regularity}. Then,
$|V_{\theta^*}^{\pi_i^*}-V_{\theta^*}^{\pi_k^*}|\leq(d_i+d_k)\normi{\theta^*-\theta}\eqsp.$ According to
Assumption~\ref{assum:zones}, one can choose $\theta$ such that $\normi{\theta^*-\theta}<c'_2\epsilon(n)$
for which $ R_n(\theta^*)\leq\E_{\theta^*}(T_n)+nc'\epsilon(n)+c_1'n\exp\{-\frac{1}{3}\log n\}\eqsp, $
where $c'=c'_2 \max_{i,k}(d_i+d_k)\eqsp.$

The maximal regret is obtained when the confidence region belongs to a frontier zone. According to
Assumptions~\ref{assum:confidence} and~\ref{assum:zones}, if $t$ satisfies $c_1 (\log n/t)^{1/2} <
c_2\epsilon(n)$ then $t\geq T_n$, with large probability. Therefore, $\E_{\theta^*}(T_n)\leq
(c_1^2\log n)/(c_2\epsilon(n))^2$. The regret is then bounded by
$$
\max_{\theta^*} R_n(\theta^*)\leq \frac{c_1^2 \log n}{c_2^2
  \epsilon^2(n)}+nc'\epsilon(n)+c_1'n\exp\{-\frac{1}{3}\log n\}\eqsp,
$$
which is minimized for $ \epsilon(n)=\left(\frac{2c_1^2\eqsp\log n}{c_2^2c'\eqsp
    n}\right)^{1/3}\eqsp.
\label{eq:epsilon(n)}
$

\section{Appendix: Proof of Theorem~\ref{theo:RegretLog}}
\label{appendix:proofTheoRegretLog}
The condition $\min_{\theta\notin Z}|\theta^*-\theta|>\kappa$ means that the distance between 
$\theta^*$ and any border of the
policy zone $Z$ is larger than $\kappa$. Hence, as soon as $\delta(\Delta_t)\leq \kappa$, the
confidence region $\Delta_t$ is included in the policy zone $Z$. The regret of the tiling algorithm
is then equal to $ R_n(\theta^*)=\E_{\theta^*}(T_n)+c_1'n\exp\{-2x\}\eqsp.  $ According to
Assumption~\ref{assum:confidence_bis}, if $t$ satisfies $c_1(x/t)^{1/2}<\kappa$ then $t\geq T_n$
with large probability. Therefore, $\E_{\theta^*}(T_n)\leq c_1x/\kappa^2$ and the regret is bounded
by $R_n(\theta^*)=\frac{c_1 x}{\kappa^2}+c_1'n\exp\{-2x\}\eqsp,$ which is minimized for $x =
\frac{\log(2c'_1n\kappa^2/c_1^2)}{2}$. For this value of $x$, we have
$R_n(\theta^*)=\frac{c_1^2}{2\kappa^2}(\log(n) + \log(2c'_1\kappa^2/c_1^2)+1)\eqsp.$

\section{Appendix: Confidence interval for Markov Chains}
\label{appendix:confidenceIntervalMC}

In this appendix, we prove that the confidence region $\Delta_t$ defined in
equation~\eqref{eq:delta} satisfies Assumption~\ref{assum:confidence}. First, remark that the event
$\{\delta(\Delta_t)\leq c_1 \frac{\sqrt{\log n}}{\sqrt{t}}\} = \{ N_t^0\geq c\frac{\eta t}{2},\eqsp
N_t^1\geq c\frac{\eta t}{2}\}$ for $c_1 = 2/\sqrt{3c\eta}$. Hence, using the Hoeffding inequality,
we have $\P_{(\alpha,\beta)}\left((\alpha,\beta) \notin \Delta_t,\eqsp \delta(\Delta_t)\leq c_1
  \frac{\sqrt{\log n}}{\sqrt{t}}\right) \leq 4\exp\{-\frac{1}{3}\log n\}\eqsp.  $ Moreover, we need
to bound the probability 
$\P\left(\delta(\Delta_t)> c_1 \frac{\sqrt{\log n}}{\sqrt{t}}\right)$. We apply Theorem 2 of
\cite{Glynn:Ormoneit:02} to bound $\P\left(N_t^1<c\frac{\eta t}{2}\right)$.  To do so, remark that
$\inf_{\alpha,\beta}\nu_1=\eta$ and that the minoration constant $1-|\beta-\alpha|$ is lower-bounded
by $2\eta$. We then have
$$
\P\left(N_t^1<c\frac{\eta t}{2}\right)\leq\P\left(N_t^1-\nu_1t<-(1-c/2)\nu_1t\right)\leq
\exp\{-\frac{4\eta^2(t^2\eta(1-c/2)-1/\eta)^2}{2t}\}\leq \exp\{-\frac{1}{3}\log(n)\} \eqsp,
$$
where the last inequality holds for $t\geq t_n\eqdef
(8/3\log(n)\eta^{-4}(2-c)^{-2})^{1/3}$. Similarly, we can show that, for $t\geq t_n$,
$\P(N_t^0<c\frac{\eta t}{2})\leq \exp\{-\frac{1}{3}\log(n)\}\eqsp.$ Hence, for all $t\geq t_n$,
$\P\left(\delta(\Delta_t)> c_1 \frac{\sqrt{\log n}}{\sqrt{t}}\right)\leq
2\exp\{-\frac{1}{3}\log(n)\}\eqsp.$ In addition, for all $t< t_n$, $c_1\sqrt\frac{\log n}{t}>
c_1\sqrt\frac{\log n}{t_n}\geq 1\eqsp,$ for $n\geq \exp\{3\times
2^{-3/2}c^{3/2}(2-c)^{-1}\eta^{-1/2}\}\eqdef n_{\min}$.  Then, for $t< t_n$ and $n\geq n_{\min}$,
the event $\{\delta(\Delta_t)\leq c_1 \frac{\sqrt{\log n}}{\sqrt{t}}\}$ is always
verified. To conclude, we have
\begin{align*}
  &\P_{(\alpha,\beta)}\left((\alpha,\beta) \in \Delta_t,\eqsp \delta(\Delta_t)\leq c_1 \frac{\sqrt{\log n}}{\sqrt{t}}\right)\\
  &\qquad\geq 1-\P_{(\alpha,\beta)}\left(\delta(\Delta_t)> c_1 \frac{\sqrt{\log n}}{\sqrt{t}}\right) - \P_{(\alpha,\beta)}\left((\alpha,\beta) \notin \Delta_t,\eqsp \delta(\Delta_t)\leq c_1 \frac{\sqrt{\log n}}{\sqrt{t}}\right)\geq1-6\exp\{-\frac{1}{3}\log(n)\}\eqsp.\\
\end{align*}

\bibliographystyle{abbrv}
\bibliography{biblio}

\end{document}

%% file: channel1ter.pstex_t
\begin{picture}(0,0)%
\includegraphics{channel1ter.pstex}%
\end{picture}%
\setlength{\unitlength}{2486sp}%
\begingroup\makeatletter\ifx\SetFigFontNFSS\undefined%
\gdef\SetFigFontNFSS#1#2#3#4#5{%
  \reset@font\fontsize{#1}{#2pt}%
  \fontfamily{#3}\fontseries{#4}\fontshape{#5}%
  \selectfont}%
\fi\endgroup%
\begin{picture}(7651,3627)(-599,-2776)
\put(1546,-2536){\makebox(0,0)[lb]{\smash{{\SetFigFontNFSS{7}{8.4}{\rmdefault}{\mddefault}{\updefault}{\color[rgb]{0,0,0}Slot}%
}}}}
\put(1658,-2761){\makebox(0,0)[lb]{\smash{{\SetFigFontNFSS{7}{8.4}{\rmdefault}{\mddefault}{\updefault}{\color[rgb]{0,0,0}1}%
}}}}
\put(1711,-736){\makebox(0,0)[b]{\smash{{\SetFigFontNFSS{7}{8.4}{\rmdefault}{\mddefault}{\updefault}{\color[rgb]{0,0,0}$X_1(2)=1$}%
}}}}
\put(1711,-2086){\makebox(0,0)[b]{\smash{{\SetFigFontNFSS{7}{8.4}{\rmdefault}{\mddefault}{\updefault}{\color[rgb]{0,0,0}$X_1(N)=0$}%
}}}}
\put(1711,164){\makebox(0,0)[b]{\smash{{\SetFigFontNFSS{7}{8.4}{\rmdefault}{\mddefault}{\updefault}{\color[rgb]{0,0,0}$X_1(1)=0$}%
}}}}
\put(2671,-2536){\makebox(0,0)[lb]{\smash{{\SetFigFontNFSS{7}{8.4}{\rmdefault}{\mddefault}{\updefault}{\color[rgb]{0,0,0}Slot}%
}}}}
\put(2783,-2761){\makebox(0,0)[lb]{\smash{{\SetFigFontNFSS{7}{8.4}{\rmdefault}{\mddefault}{\updefault}{\color[rgb]{0,0,0}2}%
}}}}
\put(2836,164){\makebox(0,0)[b]{\smash{{\SetFigFontNFSS{7}{8.4}{\rmdefault}{\mddefault}{\updefault}{\color[rgb]{0,0,0}$X_2(1)=1$}%
}}}}
\put(2836,-736){\makebox(0,0)[b]{\smash{{\SetFigFontNFSS{7}{8.4}{\rmdefault}{\mddefault}{\updefault}{\color[rgb]{0,0,0}$X_2(2)=0$}%
}}}}
\put(2836,-2086){\makebox(0,0)[b]{\smash{{\SetFigFontNFSS{7}{8.4}{\rmdefault}{\mddefault}{\updefault}{\color[rgb]{0,0,0}$X_2(N)=0$}%
}}}}
\put(3796,-2536){\makebox(0,0)[lb]{\smash{{\SetFigFontNFSS{7}{8.4}{\rmdefault}{\mddefault}{\updefault}{\color[rgb]{0,0,0}Slot}%
}}}}
\put(3908,-2761){\makebox(0,0)[lb]{\smash{{\SetFigFontNFSS{7}{8.4}{\rmdefault}{\mddefault}{\updefault}{\color[rgb]{0,0,0}3}%
}}}}
\put(3961,164){\makebox(0,0)[b]{\smash{{\SetFigFontNFSS{7}{8.4}{\rmdefault}{\mddefault}{\updefault}{\color[rgb]{0,0,0}$X_3(1)=0$}%
}}}}
\put(3961,-736){\makebox(0,0)[b]{\smash{{\SetFigFontNFSS{7}{8.4}{\rmdefault}{\mddefault}{\updefault}{\color[rgb]{0,0,0}$X_3(2)=1$}%
}}}}
\put(3961,-2086){\makebox(0,0)[b]{\smash{{\SetFigFontNFSS{7}{8.4}{\rmdefault}{\mddefault}{\updefault}{\color[rgb]{0,0,0}$X_3(N)=1$}%
}}}}
\put(4898,-2536){\makebox(0,0)[lb]{\smash{{\SetFigFontNFSS{7}{8.4}{\rmdefault}{\mddefault}{\updefault}{\color[rgb]{0,0,0}Slot}%
}}}}
\put(5011,-2761){\makebox(0,0)[lb]{\smash{{\SetFigFontNFSS{7}{8.4}{\rmdefault}{\mddefault}{\updefault}{\color[rgb]{0,0,0}4}%
}}}}
\put(5063,164){\makebox(0,0)[b]{\smash{{\SetFigFontNFSS{7}{8.4}{\rmdefault}{\mddefault}{\updefault}{\color[rgb]{0,0,0}$X_4(1)=0$}%
}}}}
\put(5063,-736){\makebox(0,0)[b]{\smash{{\SetFigFontNFSS{7}{8.4}{\rmdefault}{\mddefault}{\updefault}{\color[rgb]{0,0,0}$X_4(2)=0$}%
}}}}
\put(5063,-2086){\makebox(0,0)[b]{\smash{{\SetFigFontNFSS{7}{8.4}{\rmdefault}{\mddefault}{\updefault}{\color[rgb]{0,0,0}$X_4(N)=0$}%
}}}}
\put(6046,-2536){\makebox(0,0)[lb]{\smash{{\SetFigFontNFSS{7}{8.4}{\rmdefault}{\mddefault}{\updefault}{\color[rgb]{0,0,0}Slot}%
}}}}
\put(6158,-2761){\makebox(0,0)[lb]{\smash{{\SetFigFontNFSS{7}{8.4}{\rmdefault}{\mddefault}{\updefault}{\color[rgb]{0,0,0}5}%
}}}}
\put(6211,164){\makebox(0,0)[b]{\smash{{\SetFigFontNFSS{7}{8.4}{\rmdefault}{\mddefault}{\updefault}{\color[rgb]{0,0,0}$X_5(1)=1$}%
}}}}
\put(6211,-736){\makebox(0,0)[b]{\smash{{\SetFigFontNFSS{7}{8.4}{\rmdefault}{\mddefault}{\updefault}{\color[rgb]{0,0,0}$X_5(2)=1$}%
}}}}
\put(6211,-2086){\makebox(0,0)[b]{\smash{{\SetFigFontNFSS{7}{8.4}{\rmdefault}{\mddefault}{\updefault}{\color[rgb]{0,0,0}$X_5(N)=0$}%
}}}}
\put(6976,164){\makebox(0,0)[lb]{\smash{{\SetFigFontNFSS{7}{8.4}{\rmdefault}{\mddefault}{\updefault}{\color[rgb]{0,0,0}t}%
}}}}
\put(6976,-736){\makebox(0,0)[lb]{\smash{{\SetFigFontNFSS{7}{8.4}{\rmdefault}{\mddefault}{\updefault}{\color[rgb]{0,0,0}t}%
}}}}
\put(-539,389){\makebox(0,0)[lb]{\smash{{\SetFigFontNFSS{7}{8.4}{\rmdefault}{\mddefault}{\updefault}{\color[rgb]{0,0,0}bandwidth: B(1)}%
}}}}
\put(-584,-1861){\makebox(0,0)[lb]{\smash{{\SetFigFontNFSS{7}{8.4}{\rmdefault}{\mddefault}{\updefault}{\color[rgb]{0,0,0}bandwidth: B(N)}%
}}}}
\put(-404,614){\makebox(0,0)[lb]{\smash{{\SetFigFontNFSS{8}{9.6}{\rmdefault}{\mddefault}{\updefault}{\color[rgb]{0,0,0}Channel 1}%
}}}}
\put(-449,-1636){\makebox(0,0)[lb]{\smash{{\SetFigFontNFSS{8}{9.6}{\rmdefault}{\mddefault}{\updefault}{\color[rgb]{0,0,0}Channel N}%
}}}}
\put(-404,-286){\makebox(0,0)[lb]{\smash{{\SetFigFontNFSS{8}{9.6}{\rmdefault}{\mddefault}{\updefault}{\color[rgb]{0,0,0}Channel 2}%
}}}}
\put(-539,-511){\makebox(0,0)[lb]{\smash{{\SetFigFontNFSS{7}{8.4}{\rmdefault}{\mddefault}{\updefault}{\color[rgb]{0,0,0}bandwidth: B(2)}%
}}}}
\put(6931,-2086){\makebox(0,0)[lb]{\smash{{\SetFigFontNFSS{7}{8.4}{\rmdefault}{\mddefault}{\updefault}{\color[rgb]{0,0,0}t}%
}}}}
\end{picture}%

%% file: transition.pstex_t
\begin{picture}(0,0)%
\includegraphics{transition.pstex}%
\end{picture}%
\setlength{\unitlength}{3108sp}%
\begingroup\makeatletter\ifx\SetFigFontNFSS\undefined%
\gdef\SetFigFontNFSS#1#2#3#4#5{%
  \reset@font\fontsize{#1}{#2pt}%
  \fontfamily{#3}\fontseries{#4}\fontshape{#5}%
  \selectfont}%
\fi\endgroup%
\begin{picture}(2904,968)(-8,-69)
\put(2881,164){\makebox(0,0)[b]{\smash{{\SetFigFontNFSS{11}{13.2}{\rmdefault}{\mddefault}{\updefault}{\color[rgb]{0,0,0}$\beta_i$}%
}}}}
\put(1171,704){\makebox(0,0)[b]{\smash{{\SetFigFontNFSS{11}{13.2}{\rmdefault}{\mddefault}{\updefault}{\color[rgb]{0,0,0}$\alpha_i$}%
}}}}
\end{picture}%

%% file: TilingAlgoForChannels_hal.bbl
\begin{thebibliography}{10}

\bibitem{akyildiz:won-yeol:vuran:mohanty:2008}
I.~F. Akyildiz, L.~Won-Yeol, M.~C. Vuran, and S.~Mohanty.
\newblock A survey on spectrum management in cognitive radio networks.
\newblock {\em IEEE Communications Magazine}, 46(4):40--48, 2008.

\bibitem{Auer:Ortner:07}
P.~Auer and R.~Ortner.
\newblock {Logarithmic online regret bounds for undiscounted reinforcement
  learning}.
\newblock {\em Advances in Neural Information Processing Systems: Proceedings
  of the 2006 Conference}, page~49, 2007.

\bibitem{Glynn:Ormoneit:02}
P.~Glynn and D.~Ormoneit.
\newblock Hoeffding's inequality for uniformly ergodic {M}arkov chains.
\newblock {\em Statistics and Probability Letters}, 56(2):143--146, 2002.

\bibitem{Guha:Munagala:07}
S.~Guha and K.~Munagala.
\newblock Approximation algorithms for partial-information based stochastic
  control with {M}arkovian rewards.
\newblock {\em Foundations of Computer Science, 2007. FOCS'07. 48th Annual IEEE
  Symposium on}, pages 483--493, 2007.

\bibitem{haykin:2005}
S.~Haykin.
\newblock Cognitive radio: Brain-empowered wireless communications.
\newblock {\em IEEE J. Selected Areas Commun.}, 23(2):201--220, 2005.

\bibitem{Lai:al:08}
L.~Lai, H.~El~Gamal, H.~Jiang, and H.~Vicent~Poor.
\newblock Optimal medium access protocols for cognitive radio networks.
\newblock In {\em 6th International Symposium on Modeling and Optimization in
  Mobile, Ad Hoc, and Wireless Networks and Workshops}, 2008.

\bibitem{LeNy:al:08bis}
J.~Le~Ny, M.~Dahleh, and E.~Feron.
\newblock Multi-{UAV} dynamic routing with partial observations using restless
  bandit allocation indices.
\newblock In {\em American Control Conference, 2008}, pages 4220--4225, 2008.

\bibitem{Liu:Zhao:08}
K.~Liu and Q.~Zhao.
\newblock A restless bandit formulation of opportunistic access: Indexablity
  and index policy.
\newblock {\em 5th IEEE Annual Communications Society Conference on Sensor,
  Mesh and Ad Hoc Communications and Networks Workshops, 2008. SECON Workshops'
  08}, pages 1--5, 2008.

\bibitem{Long:al:08}
X.~Long, X.~Gan, Y.~Xu, J.~Liu, and M.~Tao.
\newblock An estimation algorithm of channel state transition probabilities for
  cognitive radio systems.
\newblock In {\em Cognitive Radio Oriented Wireless Networks and
  Communications}, 2008.

\bibitem{mitola:2000}
J.~Mitola.
\newblock {\em Cognitive Radio - An Integrated Agent Architecture for Software
  Defined Radio}.
\newblock PhD thesis, Royal Institute of Technology, Kista, Sweden, May 8 2000.

\bibitem{Papadimitriou:Tsitsiklis:94}
C.~Papadimitriou and J.~Tsitsiklis.
\newblock The complexity of optimal queueing network control.
\newblock {\em Structure in Complexity Theory Conference, 1994., Proceedings of
  the Ninth Annual}, pages 318--322, 1994.

\bibitem{Strehl:Littman:08}
A.~Strehl and M.~Littman.
\newblock An analysis of model-based interval estimation for {M}arkov decision
  processes.
\newblock {\em Journal of Computer and System Sciences}, 74(8):1309--1331,
  2008.

\bibitem{Sutton:92}
R.~Sutton.
\newblock {\em Reinforcement Learning}.
\newblock Springer, 1992.

\bibitem{Tewari:Bartlett:08}
A.~Tewari and P.~Bartlett.
\newblock Optimistic linear programming gives logarithmic regret for
  irreducible {MDPs}.
\newblock {\em Advances in Neural Information Processing Systems},
  20:1505--1512, 2008.

\bibitem{Whittle:88}
P.~Whittle.
\newblock Restless bandits: Activity allocation in a changing world.
\newblock {\em Journal of Applied Probability}, 25:287--298, 1988.

\bibitem{Zhao:al:08}
Q.~Zhao, B.~Krishnamachari, K.~Liu, M.~McKay, P.~Smith, H.~Suraweera,
  I.~Collings, Y.~Reznik, G.~Champenois, G.~Khodak, et~al.
\newblock On myopic sensing for multi-channel opportunistic access: Structure,
  optimality, and performance.
\newblock {\em IEEE Trans. Wireless Communications}, 7:5431--5440, 2008.

\bibitem{Zhao:al:07bis}
Q.~Zhao, L.~Tong, A.~Swami, and Y.~Chen.
\newblock Decentralized cognitive {MAC} for opportunistic spectrum access in ad
  hoc networks: A {POMDP} framework.
\newblock {\em IEEE Journal on Selected Areas in Communications},
  25(3):589--600, 2007.

\end{thebibliography}
